# Why we need biased AI
## How including cognitive and ethical machine biases can enhance AI systems

Sarah Fabi
sarah.fabi@uni-tuebingen.de
University of Tuebingen
Neuro-Cognitive Modeling Group

Thilo Hagendorff[*]
thilo.hagendorff@uni-tuebingen.de
University of Tuebingen
Cluster of Excellence "Machine Learning: New Perspectives for Science"

**Abstract** – This paper stresses the importance of biases in the field of artificial intelligence (AI) in two regards. First, in order to foster efficient algorithmic decision-making in complex, unstable, and uncertain real-world environments, we argue for the structurewise implementation of human cognitive biases in learning algorithms. Secondly, we argue that in order to achieve ethical machine behavior, filter mechanisms have to be applied for selecting biased training stimuli that represent social or behavioral traits that are ethically desirable. We use insights from cognitive science as well as ethics and apply them to the AI field, combining theoretical considerations with seven case studies depicting tangible bias implementation scenarios. Ultimately, this paper is the first tentative step to explicitly pursue the idea of a re-evaluation of the ethical significance of machine biases, as well as putting the idea forth to implement cognitive biases into machines.

## 1 Introduction

The overarching, common denominator for our compilation of arguments for bias implementation lies in the requirement that machine learning systems need intentional interventions to entrench presets, tendencies, or imbalances towards particular values, or targeted restrictions of the stimuli they are processing. What does that mean? Often, machine learning applications are supposed to be trained on as much data as possible, on data from whole populations, that is, on data that represents "everyone and everything". Furthermore, machine learning models, concerning their technical interior, often work via tabula-rasa-like, value-free, indeterminate neural artificial networks that lack prior "knowledge", that tweak themselves automatically by being "fed" with data and by "blindly" making associations and detecting correlations. In order to circumvent this, more inspiration from human cognition as well as intervention to create data biases becomes necessary. To intentionally include ethical data biases goes against the mainstream discourse in the field, but we think that it is important to reevaluate the meaning and importance of intentional bias promotion and implementation. The idea to include cognitive biases into machine learning algorithms has, at least to our knowledge, not been raised in the literature so far, since human biases have long been seen as violations of rationality standards, as limitations to intelligence, or simply as flaws. Nowadays, a more nuanced and positive view of human cognitive biases has been established that leads to the idea of including those into machines. We argue for a re-evaluation of the notion of ethical and cognitive biases in machines, which can be ethically desirable as well as methodologically advantageous when implemented in machine learning models.

---

[*] Corresponding author

The paper consists of three main chapters. The first chapter outlines various types of machine biases and builds the scaffolding for the differentiation of the successive two chapters on cognitive as well as ethical machine biases. In the second chapter on cognitive machine biases, we primarily regard human cognitive biases and discuss potential advantages of replicating them in learning algorithms. We argue that in unstable, complex real-world scenarios, where the future can differ significantly from the past, algorithm biases can help to learn faster and to have more accurate algorithmic decision-making as well as generalizability. We use the psychological debate on human cognitive biases and transfer it to considerations on model architectures. In the third chapter on ethical machine bias, we reintroduce the idea of algorithmic discrimination in an altered, but positive manner. We stress that in order to achieve ethical machine behavior one must not debias data sets. The mere notion of "debiasing" machine learning applications evokes a problematic myth of "neutrality", which is not only unachievable in the sense of perfect fairness (as the impossibility theorem proves (Kleinberg et al. 2016)), but which also masks that neutrality can be a contestable value or political stance. We argue that instead of debiasing data sets, one should systematically "distort" selection processes of data sources and intentionally prioritize certain features over others, namely features that are desirable from an ethical point of view. Moreover, each of the chapters on bias implementation contains case studies including tentative examples that amend our theoretical considerations with practical application scenarios.

In general, this paper breaks new ground. The considerations we are presenting are still in the early stages of maturation. This paper is the first to explore the idea that biased AI can actually be important from a cognitive and ethical science perspective. The chapters are rather comments, tentative explorations of a new field, sometimes speculative first steps in a direction where we think that the machine learning community has blind spots or can make improvements by reassessing the value of biases. We want to encourage the community to further investigate the potential importance of ethical and cognitive machine biases and to conceive model architectures that technically implement bias-like structures. This paper, however, does not contain details on technical frameworks. Instead, in a first step, we wanted to map out arguments on the actual importance of different kinds of biases in AI systems and invite the machine learning community to consider reevaluating machine biases in a more nuanced way.

## 2   Types of machine biases

For our paper, we describe in detail the importance of cognitive as well as ethical machine biases in AI applications. But first of all, we want to clarify terms. Over the last decades, "bias" became a term riddled with ambiguities, especially if one looks into different scientific disciplines and fields. However, we think that a common denominator for all types of biases we elaborate on is that they can be paraphrased as some kind of distortion, as a tendency towards a particular value, as a specific presetting, or simply as deviation from a standard (Danks and London 2017). Despite that, we argue that, for the purpose of this work, three different denotations of the term "bias" can be separated from each other, namely machine biases in the fairness field, machine learning specific inductive biases, and human cognitive biases.

First of all, machine biases in the fairness field are associated with algorithmic discrimination, which, roughly speaking, stands for disparate, unjust impacts of applications of algorithmic decision making on individuals (Barocas and Selbst 2016; Fabi et al. 2022b). Respective fairness machine biases can be entrenched in machine learning applications via data, human-computer interactions, as well as algorithms (Hellström et al. 2020; Mehrabi et al. 2019). First, data bias is a systematic distortion in the sampled data that can be caused by selection processes of data sources, the way in which data from these sources are acquired, as well as by processing operations such as cleaning or aggregation (Olteanu et al. 2019). Akin to data biases are human-computer interaction biases. Human-computer interactions can be affected by specific behavioral patterns in humans, ultimately affecting the very data that is used for further model training. Secondly, algorithms can become biased by human preferences that influence the way they are assembled, or by technical biases that are intentionally implemented or that emerge during model development, evaluation, or deployment (Suresh and Guttag 2020). In all these areas, various more specific types of distinct biases can be differentiated. However, the different types are not clearly distinct from each other. Quite the contrary, the differentiations that are in use contain various overlaps and intersections between the biases. All in all, over the last years, the term "bias" became synonymous with all kinds of unjust machine behavior in the fairness field. Machine learning applications, with their focus on large training data sets and accuracy towards mean values of whole populations, must shy away from distortions that are the result of machine biases. But as soon as these applications are seen as systems that shift from the big data rationale *n = all* to a more selective way of processing data for training sets for ethical reasons (Hagendorff 2021c), putting particular ethically motivated machine biases in place can prove to be very helpful.

Inductive biases, on the other hand, are mainly used in the technical field of machine learning and are defined as priors or assumptions of an algorithm to build a general model out of a limited set of training data (Hüllermeier et al. 2013). Such biases are necessary as there exist too many possible ways to perform this induction (Mitchell 1980). In other words, inductive biases are constraints of the algorithm's hypothesis space (Baxter 2000). This hypothesis space should, on the one hand, be small enough to ensure generalization and, on the other hand, large enough so that a solution to the learning task can be found. Most often, the hypothesis space is selected by the experimenter designing the learning algorithm. It becomes clear that inductive biases are necessary for the success of learning algorithms because unbiased algorithms make no a priori assumptions about probabilities of predictions and instead rely solely on the presented training data, making successful generalization impossible (Mitchell 1980). However, the "no free lunch theorem" shows that one cannot define an inductive bias about what functions or classifications to use that is suited best for a given out-of-distribution generalization task (Wolpert 1996; Wolpert and Macready 1997). Examples are classification algorithms that are biased to assume neighboring data points to belong to the same group, linear models that impute linear boundaries between the groups, LSTM models that are biased to memorize information over time, Bayesian models with specific prior distributions, and many more (Goyal and Bengio 2021). Other inductive biases are inspired by human learning dispositions, for example the fact that humans compose objects into parts that they can reassemble to learn more efficiently (Fabi et al. 2021b; Fabi et al. 2020; Fabi et al. 2022a; Fabi et al. 2021a). Sometimes, inductive biases are used synonymously with priors or learning biases.

Whereas fairness and inductive biases represent biases in machine learning systems, cognitive biases stand for human biases that are scrutinized with means of psychology and that can be due to intelligence limitations, motivational factors, and adaptations to situational circumstance (Tversky and Kahneman 1974). Cognitive biases are defined as gaps between expected, rational behavior on the one hand and heuristically determined behavior on the other hand. Heuristics are shortcuts that reduce task complexity in judgments and choices. Among others, human cognitive biases can take effect during collecting empirical knowledge, when assessing evidence, or when gauging the likelihood of events. They have implications for learning processes, for instance when probabilities are overestimated (conjunction fallacy), prior probabilities are neglected (base-rate neglect), rules that confirm prior hypothesis are "cherry picked" (confirmation bias), information that is presented first has the highest impact (primacy effect), one's own cognitive biases are not compensated for (bias blind spot), and many more (Kliegr et al. 2021). In general, biases help to simplify decision making processes. Moreover, they allow decision making in view of large amounts of data and complex interdependencies in the data in the first place. In the next chapter, we want to combine considerations on inductive biases and human cognitive biases by transferring the latter into the former. Machine learning applications that operate close-end classification problems, possess nearly unlimited computational resources for training, and application scenarios that are devoid of significant novelties and changes can presumably not profit from technically imitating human cognitive biases. But as soon as these applications have to manage uncertain, complex, and rapidly changing environments where the distribution of observations may change continuously, putting particular machine biases that are inspired by human cognitive biases in place can again prove to be very helpful and foster efficient learning of strong out-of-distribution generalization capabilities and transfer learning.

In the following, we stress technical as well as ethical rationales for bias implementation in machine learning systems. While ethical demands are aiming strongly for a socially acceptable deployment of machine learning applications, eventually leading to better machine behavior (Rahwan et al. 2019) that circumvents a lot of the socially relevant pitfalls and normative shortcomings current machine learning systems fall victim to, technical advancements aim at increasing learning algorithm's performance. Hence, the following two chapters are both collections of arguments and tentative ideas for bias implementation.

## 3   Cognitive machine biases

Cognitive machine bias can be a result of aims to technically advance machine learning architectures via particular inductive biases. The idea that it is promising to include inductive biases into learning algorithms has become quite acknowledged in recent years (Battaglia et al. 2018; Goyal and Bengio 2021). What is special, in our case, is that we request those inductive biases to be replicating human cognitive biases. Hence, when we use the term cognitive machine bias here, we are focusing on potential ways to replicate high-level cognitive biases in learning algorithms. But why can cognitive machine biases in this sense be important in the first place? Human intelligence has three major restrictions that artificial intelligence and its theoretical "unbounded rationality" have not (Griffiths 2020): It has a limited amount of time and thus limited data for problem-solving, it has a limited amount of computational resources, meaning resource limitations that are reflected through anatomical, physiological, and metabolic constraints (Lieder and Griffiths 2019), and it has limited possibilities for communication, meaning that humans cannot directly transfer information from one brain to another. Understanding how human intelligence navigates these limitations can help inspire efficient problem-solving and learning methods in future machines. Current machine learning applications often simply increase the amount of computation or the amount of training data that is used to solve problems. GPT-3, to name a well-known example, has been trained on so many text tokens that when trying to produce them all, one would have to speak continuously for 5.000 years (Griffiths 2020). It is the missing limitations and the missing corresponding evolutionary pressure that led to artificial intelligence systems that do not show the traits associated with human intelligence, like efficient learning, the development of structured representations, meta-reasoning etc. (Griffiths et al. 2019).

However, even if limited resources led to efficient learning during the evolution of humans, just limiting resources for algorithms will not solve the problem. Thus, the demand to include inductive biases into machine learning algorithms became more and more apparent in the last years (Amatriain 2019; Marcus 2018). In this paper, in accordance to Goyal and Bengio who argue "that having larger and more diverse datasets is important but insufficient without good architectural inductive biases" (2021), we want to outline ideas on how to include priors in the form of human cognitive biases into algorithms in order to qualitatively progress learning methods as cognitive biases constitute one way for humans to deal with their limitations. In other words, especially when presented with fewer training data, we expect that reverse-engineering human cognitive biases might help machine learning systems to work more efficiently and accurately, among other advantages, which we will discuss in more detail. Thus, in this chapter, we explain why it will be helpful to include human cognitive biases into learning machines. To understand what advantages this inclusion can bring and in which ways cognitive biases have already been implemented in machine learning algorithms implicitly and explicitly, we first have to look into the psychological debate of human cognitive biases.

### 3.1 Human cognitive biases as flaws vs. essential tools for complex, real-world decision making

Humans have several systematic misconceptions, insensitivities to chances, systematic errors, etc. In general, they violate particular statistical rules, revealing that correct statistical notions are not part of their repertoire of intuitions. Examples for typical cognitive biases are the confirmation bias where individuals have a tendency to pay more attention to information that confirm existing beliefs, the anchoring bias where humans are overly influenced by information they hear first, or the availability bias that causes people to overestimate the probability of an event as long as many examples of the very event readily come to mind. Already in 1974, behavioral economists found that decision-making under uncertainty is faulty since humans rely on a limited number of heuristic principles (Tversky and Kahneman 1974). These principles reduce complex decision-making tasks to simple and error-prone judgmental operations. It is noteworthy, though, that a predominantly negative view of human cognitive biases gradually made way for a more positive concept over time. Cognitive mechanisms in intuitive statistics or probabilistic reasoning might even be necessary prerequisites to make sense of a highly complex world. This means that in order to intelligently and effectively navigate and interact with complex environments in unfamiliar and uncertain situations with limited amounts of information, time, and energy resources, humans need heuristics, meaning mental shortcuts that they use to make quick, intuitive decisions. Heuristics are acquired through evolution and experience and are based on the environmental regularities in which humans routinely interact (Korteling et al. 2018). According to the "evolutionary" or "ecological perspective" (Gigerenzer 1991, 1996; Gigerenzer and Gaissmaier 2011; Haselton et al. 2009; Vranas 2000), only under conditions of a mismatch between current environments and evolutionary adaptation, in artificial laboratory settings, in particular cultural transformations of "natural" lifeworlds, or when lacking relevant expertise, biases become problematic flaws. Deviations from tenets of rationality or violations of rules of logic and probability are obviously not always useful. They can without a doubt lead to suboptimal decision making in many areas. However, cognitive biases are an essential part of naturalistic decision-making (Klein 2008). Eventually, when considering complex real-world environments, human biases and heuristics can be turned from a phenomenon that has to be avoided into a tool that might improve decisions, which is transferable in a certain way to artificial learning systems (Gadzinski and Castello 2020). Therefore, in the following, we will identify advantages that might result from supplying learning machines with cognitive biases. We will discuss in more detail the advantages of cognitive biases in human decision-making and probe whether these advantages can be transferred to algorithms.

## 3.2 Do advantages of human cognitive biases hold for machines, too?

The most prominent argument for the use of cognitive biases in humans is their effort reduction and speed. Humans need biases because they do not have enough computational power to take all the available information of their environment into account (Payne et al. 2012). Of course, the limited computational resources of humans are not present in computers. But apart from considerations on evolutionary pressure with regard to intelligence formation mentioned at the beginning of the chapter and considerations regarding "Green AI", meaning resource efficiency and less computational requirements in machine learning systems (Schwartz et al. 2019; Makridakis et al. 2018; Patterson et al. 2021; Strubell et al. 2019), Gigerenzer and his group (Hafenbrädl et al. 2016) state that effort reduction is just a by-product in humans' use of heuristics. More important than effort reduction is the improved accuracy that comes with cognitive heuristics.

In situations of high uncertainty, taking all the available information into account does not lead to better, but sometimes even worse predictions and decisions. There is no such thing as a trade-off between accuracy and information reduction or bounded reasoning. In fact, as psychological research shows (Gigerenzer and Gaissmaier 2011), simple heuristics that use the most relevant information outperform more complex reasoning techniques that use large amounts of information. This holds true for machines as well, for instance, when predicting occurrences of infectious diseases (Katsikopoulos et al. 2022). This means that research from other disciplines than psychology proves as well that more complex models for decision-making under uncertainty can increase error rates compared to simple procedures (Green and Armstrong 2015). All in all, one cannot state that complex machine learning models categorically form better predictions than simple ones. Especially in forecasting tasks, and thus insecure environments, their superiority is not apparent (Makridakis et al. 2018, 2020; Smyl 2020). This is why implementing cognitive biases into model architectures might lead to a simplification, to a complexity reduction in how the very models process training stimuli, but it is their advance in terms of forecasting performance and accuracy.

But how can more complex decisional methods under conditions of uncertainty lead to less accurate predictions? Especially in situations in which the future differs significantly from the past, biases can help to adapt to changes faster than complex machine learning models that inflexibly rely on large amounts of past data. Moreover, a paradigm known in psychology as well as in the machine learning community is the bias-variance dilemma (Geman et al. 1992; Gigerenzer and Brighton 2009; Luxburg and Schoelkopf 2008). When humans make decisions, they generate models of their environment. The goal of such models is to fit the available data in a way that they can form accurate predictions about unknown (future) data, also called generalizability (Pitt et al. 2002). This generalization can be impaired by two factors: If there exists a bias, the model consistently predicts deviations from the true data, e.g, by assuming a linear instead of the true quadratic relation. But it is not the bias per se that determines the accuracy of a model, but the trade-off between bias and variance. If there is too much variance in the model, it might fit very well to the presented data but cannot generalize to similar but new data. It represents not only the signal but also the overlaying noise. This can be the case if models become too complex. We claim that in complex, uncertain environments, in which the problem of generalizability of models with too many free parameters gets even harder, one antidote is to apply cognitive biases. In line with our view, Gadzinski and Castello (2020) state that heuristics minimize the variance component in the bias-variance dilemma, leading to a prediction error that is mainly defined by the bias, enhancing the ability to fit the data. According to Katsikopoulos et al. (2022), in certain environments, the bias of a heuristic is even equal to the bias of an optimal linear model.

In machine learning algorithms, similar problems arise. If the model is too complex or too little data is available, not only the signal but also the noise is represented. Generalization gets very hard and the model is overfitting. When assessing the success of machine learning models by taking the environment or, in other words, domain knowledge into account as suggested by the theory of ecological rationality (Todd and Gigerenzer 2012), models which are prone to overfitting the training data fail since they are not well adapted to the test environment. Hence, in prediction contests like the M4 competition, pure machine learning models show poor forecasting performances (Makridakis et al. 2018, 2020). Applying human cognitive biases in certain data environments would counterbalance the high variance methodology of artificial neural nets (Gadzinski and Castello 2020). Similar to Nikolopoulos and Petropoulos (2018), who warn against too complex forecasting models that focus too much on minimizing the bias, ending up with too much variance, we assume that current machine learning algorithms are on a similar path one could avoid by implementing cognitive biases.

Next to effort reduction and increased accuracy, further advantages of human cognitive biases in psychology are accessibility and transparency (Hafenbrädl et al. 2016). Biased human decision-making is accessible, it does not require intense training and even lay people can make good decisions, for which normally experts are needed, if a good heuristic, for example, in form of a fast-and-frugal tree is provided, which is a simple decision tree for binary classification problems. This advantage can be transferred to the machine learning area because algorithms would benefit from less extensive training, especially regarding situations of limited amounts of training data, like in the context of our argument for ethical machine bias implementation. Moreover, one main disadvantage of artificial neural nets is their black box behavior. Including cognitive heuristics into such systems might help to make them more explainable if heuristics and thus specific decision rules are intentionally and explicitly included. However, it remains unclear to this date how to intermesh simple heuristics and machine learning architectures in detail in order to leverage the advantages of both approaches. This is a task for further research.

In general, we argue that the shift in psychology from viewing heuristics as purely irrational mistakes to being evolutionary as well as ecologically adaptive might help machine learning research to advance as well, by opening the doors to model heuristics and biases. But we do not want to claim that heuristics are per se better than current machine learning techniques or should always be the way to go. On the contrary, it is very important to consider in which environment a decision is made and whether it might be helpful to include certain cognitive biases into machine learning algorithms. Therefore, we want to probe whether the information we have about when heuristics are helpful in humans can again be transferred to machines. Whether heuristics are ecologically rational in humans depends on the domain of decision-making (Hafenbrädl et al. 2016). In a stable, calculable environment, for instance in games like chess or Go, complex machine learning techniques succeed when they are provided with a high amount of data. The reason for that lies in the high similarity between training sets and out-of-distribution data. But as demonstrated above, in situations in which machines have to make decisions under uncertainty, one needs to make it simple, to scale down the number of factors that influence a decision. This means that when using machine learning in dynamic, unstable real-world situations with small amounts of data, implemented heuristics can become a useful way to make better predictions, avoid overfitting, react better to quick changes, achieve robuster models, etc. Applying methods that are inspired by human cognitive biases and that trigger machine learning models to ignore particular parts of data or information does not cause them to produce more errors, but to produce outputs that are more accurate. The accuracy of heuristics is relative to the structure of an environment (Gigerenzer and Brighton 2009). The more noise in the environment, the more likely it is that heuristics will outperform complex approaches to decision making, whereas the selection of the "right" heuristic is dependent on the most relevant aspects of a problem. In humans, search rules are used to sift through cues in observations, memories, or information or to search for cues that are accessible. Stop rules determine when enough or the right cues are found in order to transition to the decision rule which ultimately selects a heuristic that results in a decision. Moreover, cognitive niches for limited sets of different heuristics are established by memories about past decision making situations (Gigerenzer and Brighton 2009). This adaptive toolbox of heuristics that humans possess can inspire the selection of the type of bias-laden machine learning architectures. The right selection of particular cognitive machine biases in a given data environment is of great importance when aiming at robust model outputs.

## 3.3 Empirical support for the superiority of simple models in certain environments

After having presented the theoretical idea of why it might be helpful to include cognitive biases into machine learning algorithms, we want to present some empirical basis for this claim. Unfortunately, until now, there does not exist a lot of work combining machine learning techniques with the implementation of cognitive biases. What already exists and what can give us some insights are comparisons between implemented heuristics and machine learning algorithms. Surprisingly, in a study by Makridakis et al. (2020), they compared 61 forecasting methods on 100,000 datasets, finding that simple models were better in predicting the data than more complex models. The researchers concluded that combinations of simple statistical and complex machine learning models might be best suited for forecasting tasks, supporting our own claim of combining machine learning with simple heuristics. Artinger, Kozodi, Wangenheim, and Gigerenzer (2018) investigated 60 datasets and found that the hiatus heuristic was more accurate than random forests or logistic regressions. Lee, Loughlin, and Lundberg (2002) found that a simple take-the-best heuristic exceeded a more complex Bayesian model in a literature search. In 30 classification tasks from medicine, sports, and economics, fast-and-frugal trees performed similarly to more complex algorithms like logistic regressions (Martignon et al. 2008). A very recent investigation of Katsikopoulos et al. (2022) shows that Google Flu Trends (Ginsberg et al. 2009) with 45 variables from 50 million Google search queries can be outperformed by the simple recency heuristic. Rafati, Joorabian, Mashhour, and Shaker (2021) show that artificial neural nets can be outperformed by simpler heuristic methods, in this case regarding solar power forecasting. Many other studies could be added (Makridakis et al. 2018), in which heuristics exceed complex (machine learning) models. Nevertheless, in this paper, we do not want to contrast the two methods or even downgrade the successes of machine learning algorithms. On the contrary, we think it is highly important to develop them further and combine them with successful heuristics inspired by human cognition.

## 3.4 Combining instead of contrasting machine learning and heuristic based approaches

To provide further support for our claim that instead of contrasting, the two approaches should be integrated, we further borrow insights from psychiatry. Since computational psychiatry for computers was introduced in 2020 by Schulz and Dayan (2020), we argue that in order to prevent machines from showing detrimental behavior that can be classified according to certain psychological disorders in humans, it makes sense to include cognitive biases. Difficulties in decision-making manifest in several psychological disorders, like anxiety, depression, and autism spectrum disorder (Bishop and Gagne 2018; Morsanyi and Byrne 2020). We hypothesize that those difficulties might also be attributable to missing cognitive heuristics that allow for fast and accurate decisions by not taking all the available information into account. For example, decision-making research in the autism spectrum disorder gives very interesting insights in this respect: Autistic persons are very good at making complex decisions, like important life questions, but they often have problems with deciding on daily-life questions due to overthinking of routine decisions (Morsanyi and Byrne 2020). Compared to healthy subjects, they score higher on the subscale of rational/effortful decision style but lower on the subscale of intuitive/experiential thinking style (Brosnan et al. 2016). In a similar vein, Morsanyi (2010) found that autistic adolescents apply fewer heuristics. Since artificial neural nets represent the rational/effortful decision style in the sense that they are taking all available information into account (Gadzinski and Castello 2020), one might draw the analogy by saying that such artificial neural nets have similar problems in uncertain situations like patients with autism spectrum disorder, since both are lacking the necessary "gut feeling". This is another reason why the idea is appealing to implement biases into machine learning architectures that hitherto use all available training stimuli.

### 3.5 Human cognitive biases in the social world

Furthermore, we want to make the claim that cognitive biases are not only helpful in decision-making but also in social contexts, which is why our argument can even go so far as to implement cognitive biases that genuinely affect social behavior in anthropomorphic systems of artificial intelligence. When using theories of multiple intelligences as a reference point (Gardner 1988), current artificial intelligence equals a form of visual-spatial as well as linguistic-verbal intelligence but lacks forms of interpersonal or social intelligence (Goleman 2006). Why the implementation of certain cognitive biases that influence social behavior is helpful here as well can be explained using economic games. In these games, biases lead to allegedly irrational behavior, too. For example, in the ultimatum game, where one player can distribute a reward between herself and another player while the latter can then decide whether she wants to accept or reject the offer, which would lead to both players receiving nothing, most humans do not act rationally regarding their direct utility (Calvillo and Burgeno 2015) but show acts of kindness, reciprocity, and fairness. Not accepting unfair offers, for example, seems flawed since to maximize one's own reward in this specific game, one should accept every offer. In an evolutionary sense, however, it might make sense to penalize group members who are acting unfairly, even though this does lower one's own direct reward (Haidt 2006). Whenever sophisticated, anthropomorphic artificial intelligence systems imitate human behavior, it can make sense to also imitate human cognitive biases. In recent research on artificial Theory of Mind, the demand for algorithms that understand human irrational behavior and unravel underlying beliefs and intentions becomes more apparent as well (Alanqary et al. 2021; Evans et al. 2015). Of course, one has to stress again that not all human biases are per se good, with some of them even decreasing fairness. Nevertheless, one can think of several scenarios, in which imitating human social biases might be beneficial, like in robot or virtual avatar interactions; in training to play certain games, where it might be helpful to simulate a competing player instead of a perfect machine; when training particular interactions, like situations of conflict or negotiations; or when modeling human behavior in order to gain psychological insights on hypotheses that cannot be tested empirically on real humans. Other than that, modeling human behavior might lead to the opportunity to combine the advantages of field experiments with high external validity and lab experiments with high internal validity. More specifically, human behavior is being tracked almost everywhere by digital devices. If researchers get access to this data, they can analyze behavior statistically (Agrawal et al. 2019). However, since they are not able to manipulate specific variables, the results are of correlational nature. If the data is, on the other hand, used to train machine learning algorithms to model human behavior, researchers should be able to experimentally manipulate specific variables and observe the model's behavior, which should mimic human behavior. First steps in this direction are made (Dezfouli et al. 2019; Schulz et al. 2019), for example, with recurrent neural nets that are trained to represent decision-making strategies used by humans.

## 3.6 Case studies

To further support the idea of implementing cognitive biases into machine learning algorithms and to give an impression on how this might be possible, the three case studies of this chapter will be differentiated in the following way: The first two case studies will identify several commonly used implicit implementations of human cognitive biases that have just not been identified as such so far, namely in the area of overfitting avoidance and shortcut learning. The third one will then describe first approaches that already try to implement human cognitive biases into machine learning algorithms in an explicit manner.

### 3.6.1 Case study 1: Implicit cognitive machine biases in overfitting avoidance

Most heuristics in humans have in common that they focus on a few important aspects of the information, while the rest is ignored (Gigerenzer and Gaissmaier 2011). For example, in the successful fast-and-frugal decision trees, some features are queried to make a decision, whereas all the other complex dependencies between the remaining features are ignored (Hafenbrädl et al. 2016). This focusing on the most relevant features of the data by ignoring the details representing noise instead of the signal is also what is important when algorithms shall avoid overfitting. As mentioned above, the less-is-more effect can be explained by the bias-variance dilemma (Gadzinski and Castello 2020). According to Gigerenzer and Brighton (2009), the amount of information that humans need to ignore in successful decision-making correlates with increasing unpredictability. An analogy can be drawn nicely for artificial neural nets, which are dealing with uncertainty: The more unpredictable the data is, that is the more training and test sets differ, the greater the problem of overfitting gets, if the algorithm has too many free parameters, and thus, becomes too specialized on the training set. As described above, we claim that cognitive biases can help to avoid overfitting not only in humans' models of the environment. To find out whether there are already cognitive biases in machines in place which are not identified as such, we will look more closely into some of the measures that are applied to avoid overfitting in algorithms.

And indeed, there exist several methods against overfitting that explicitly limit the amount of information, like removing irrelevant features of the data in advance or stopping the training early. This avoids an overspecialization and can be seen as not taking all dependencies of the data into account. This is interesting because it can be viewed as an implementation of the aforementioned point of artificially limiting time (Griffiths 2020). Approximating humans' limitation of time might indeed help to develop not too specialized, but representations of average concepts, similar to stereotypes. Another tool is regularization techniques that force the algorithm to become simpler. For artificial neural nets, dropout is a regularization technique that randomly deletes units (Srivastava et al. 2014), preventing the network from overspecializing on the presented data. Thus, lots of methods to prevent overfitting can be seen as implicit implementations of heuristics (Schaffer 1993).

Against this backdrop, one can indeed identify already implicitly implemented cognitive biases in machines. This supports our claim that machine learning research would benefit from including human biases in a more systematic manner in order to achieve more human-like learning. Until now, the measures to reduce overfitting are unsystematic implementations of heuristics, which have not been acknowledged as such so far. We assume that, for example, the selection of relevant features that are fed into the machine learning algorithms could be a good starting point to incorporate human cognitive biases, like, the take-the-best heuristic, selecting the feature that discriminates the right solution best. This heuristic is (for humans and therefore probably also for machines) most ecologically rational in environments in which there exists cue redundancy and variability in cue weights (Gigerenzer and Gaissmaier 2011).

### 3.6.2 Case study 2: Shortcut learning as an implicit cognitive machine bias

Next to the implementation of heuristics in overfitting avoidance, another bias seems to be already implicitly integrated into artificial neural nets. Geirhos et al. (2020) discuss so-called shortcut learning states, meaning that deep neural networks are often only superficially successful and fail when presented with new datasets since they learn shortcuts of the original dataset. One example is a network that learned to classify X-ray images correctly and when presented with images from a new hospital, it failed completely, since it had based its classification on a hospital-specific metal token on the scan (Zech et al. 2018). This is a problem, since the network does not learn anything about pneumonia (Geirhos et al. 2020). Nevertheless, though, shortcut learning is not purely problematic but led to many successes of classification algorithms. We assume that this could be the artificial pendant that the network develops on its own of the biological one-clever-cue heuristic, with which animals make decisions based on one helpful cue (e.g., gaze heuristic) (Shaffer et al. 2004). Depending on the environment, this heuristic can be ecologically rational. Regarding the environment of the deep neural network that fails when presented with images of a new hospital, it does really well on datasets from the same hospital. This means in its specific environment, the one-clever-cue heuristic can indeed lead to very good accuracy. This is comparable with the aforementioned flaws of heuristics that exist only under conditions of mismatch between current environments and evolutionary adaptation. In this case, the error lies with the experimenters who did not provide a sufficiently variable dataset. For the given dataset, the model learned a perfectly valid method of classification. Another example is a network that can classify cows correctly, but only if they were presented on a meadow instead of, for example, a beach (Beery et al. 2018). But if this network was trained to solve the task of classifying cows on meadows, it is perfectly adapted to this task. Its shortcut learning is of course not helpful if then presented with a new task like classifying cows on beaches, but for the original environment, its learning is ecologically rational. We assume that even humans learn shortcuts, even though probably not the same ones as algorithms. Geirhos et al. (2020) relate artificial shortcut learning to students learning superficially for a test instead of focusing on understanding. Similar to the debate in Psychology, we claim that the environment of the algorithm's behavior has to be taken into account. Regarding the students' goal of getting a good grade, shortcut learning might be totally valid and ecologically rational. One should not blame the students or the network that this is not what the experimenter aimed for. This does not mean that a more thorough understanding of algorithms instead of mere shortcut learning is not something one should aim for. On the contrary; nevertheless, the contribution of shortcut learning, and thus the one-clever-clue heuristic that the network developed on its own, to the current successes of image classification should not be underestimated.

### 3.6.3 Case study 3: Explicitly modeling cognitive biases in algorithms

To provide more evidence for our demand for the inclusion of cognitive biases into machines, in this case study, we will review some of the few available, explicit implementations of cognitive biases in machine learning systems. Hence, this case study will look at the first approaches that tried to combine cognitive biases more explicitly with machine learning algorithms, even though there are not yet many studies out there.

Gadzinski and Castello (2020) aimed at combining system 1 and system 2 thinking by combining fast-and-frugal trees with ensembles of artificial neural nets that estimated Bayesian uncertainty, respectively. With this, they were able to refine the decisions regarding loan granting made by fast-and-frugal trees in the following way: The model prediction of whether a loan was repaid was not solely dependent on exceeding a certain threshold in one variable. Instead, the prediction was leading to a certain probability of repayment when exceeding and when not reaching the threshold of certain variables. When applied in human decision-making, this procedure led to a reduction of overconfident predictions and helped humans build shortcuts while acquiring more data when necessary.

Further, Taniguchi, Sato, and Shirakawa (2018) incorporated the symmetric bias and the mutually exclusive bias into a Naive Bayes spam classifier, since both are believed to promote humans' faster learning and decision making (Shinohara et al. 2007). The symmetric bias describes the fact that humans assume that "if a, then b" automatically leads to "if b, then a". The mutually exclusive bias describes the fact that humans assume if something is called a "penguin", it cannot be a "rabbit" at the same time. Taniguchi et al. (2018) showed that by incorporating those biases, the model was able to outperform existing machine learning algorithms on small and biased spam classification datasets.

At least to our knowledge, these are the few papers that exist until now in the realm of implementing cognitive biases explicitly into machine learning algorithms. The described papers are very first, but also promising steps that should be followed up upon. To name some ideas, one could combine artificial neural nets with Fast and Frugal Trees, in that way that artificial neural nets preselect the important features, on which the decisions are made by the Fast and Frugal Trees. Preselecting the most relevant features for machine learning algorithms like regressions by simple heuristics that humans use could be another application, similar to Smyl (2020) who combined a simple exponential smoothing model (compare human recency heuristic) with a complex RNN architecture and won the M4 forecasting competition. This could work for combining machine learning algorithms that are not working quite well so far with heuristics instead of replacing one by the other (Katsikopoulos et al. 2022), like Google Flu trends and the recency heuristic by weighing the most recent information as more important. Another implementation idea of combining heuristics and machine learning algorithms is inspired by Agrawal et al. (2019): Simple models are designed based on heuristics and thus explicit rules. In an iterative process, their results are compared to the results of machine learning models and the simple models are adapted until they reach a similar performance. In this process, by continuously comparing what which model can and cannot do and by adapting the heuristic-based models, they can be inspired and improved through machine learning models which are in turn made more explainable.

## 4   Ethical machine biases

Bias in data sets is the root cause for one of the most significant risks of machine learning systems, namely algorithmic discrimination (Angwin et al. 2016; Hagendorff 2019a, 2019c, 2021a; Hagendorff et al. 2022; Kearns and Roth 2020; Myers West et al. 2019). Bias in data sets leads to "skewed" and unfair algorithmic decision-making. Scholars rightly criticize that machine learning techniques perpetuate existing biases that are, among other reasons, entrenched in historic data sets. Hence, they put tremendous effort into creating methods for reducing algorithmic discrimination. Many of these efforts are either aiming at in-processing techniques that modify learning algorithms in order to remove discrimination during model training or at post-processing techniques, where one tries to correct the results of already trained classifiers to achieve fairness. However, both techniques do not tackle the root cause for (un)ethical machine behavior, namely the very selection of features that are allowed to become training stimuli. We do not want to make a definitive judgment whether one is better off when changing the input/label data or when using unfiltered inputs and then try to fix the output. However, we think that in supervised machine learning, the effectiveness and rigor of pre-processing techniques that strictly exclude undesirable inputs in the first place are undervalued. Moreover, our approach is not just about avoiding unfairness, as is usually done via dealing with protected attributes like gender, age, ethnicity, etc.

(Dwork et al. 2011; Veale et al. 2018). It is about actively promoting prosocial attributes and goals. For that end, we argue that one should reintroduce the idea of algorithmic discrimination in an altered, positive manner. The basic idea is that in principle, data sets contain features that should be weighted stronger than others, perpetuating particular machine biases in an intentional manner. In this context, one can differentiate between "the world as it is" versus "the world as it should be" (Hellström et al. 2020). Models can be used to predict the world "as it is", which means to perpetuate random existing biases. Debiasing training data, in contrast, can lead to a modeling of the world "as it should be". Here, we also opt for using an understanding of "the world as it should be", but, instead of debiasing, by intentionally introducing bias.

### 4.1 Filtering training data

We argue for the promotion of machine biases in data sets that lead to a preferability of features that are desirable from an ethical point of view (Hagendorff 2021c). Ultimately, machine behavior can be significantly influenced by exposure to data of different origins and qualities. These qualities can be measured in technical terms, but also ethical ones. When describing new dimensions of data quality for machine learning, one can introduce ethically justified imbalances – in other words, select training data sets in a way that they reflect social or behavioral traits that are ethically desirable. The goal is to transition from a situation of "every training stimuli counts" to a situation of "only biased, ethically justified training stimuli should be allowed to shape a machine's behavior in contexts where it is important". Different qualities of human participation in model training can be distinguished, whereas biases towards "good" human influences on machines should always be embraced. The importance for machine biases comes from the ethical imperative to identify data sources that reflect behavior that is ethically sound, which in turn can be identified by scrutinizing particular states and traits of individuals and by selecting subpopulations that are deemed to be the most competent or morally versed group for a particular task. This practice of singling out competent subgroups is already established in particular domain-specific data annotation processes, for instance, in the area of medical AI, where only experts are allowed to provide labels (Irvin et al. 2019). Here, we argue for an extension of this approach beyond the medical field, i.e. for a generalization of a biased and selective human participation in processes of machine behavior conditioning, for instance when using text training data only from linguistically versed individuals or autopilot training stimuli only from decidedly safe and considerate drivers.

The current tenet is to follow a practice called "laissez-faire data collection" (Jo and Gebru 2019) or, in other words, the ideology of $n = all$ (Mayer-Schönberger and Cukier 2013; Perrons and McAuley 2015), in order to aim at presumably higher grades of accuracy in clustering, regression, or classification tasks. Thus, algorithmic decision-making becomes indifferent with respect to the orientation towards certain moral values. Hitherto, the overwhelming majority of machine behavior that results from recognizing statistical patterns in social data is trained by learning from general populations, not specific subgroups. As said before, these subgroups could comprise social data from individuals that are most competent, eligible, or morally versed for a particular task that is supposed to be automatized. In other words, we propose to introduce filters that thwart particular data traces to become training data for machine learning. In particular, we argue for representational as well as population bias in training data sets. This way, inputs for model training manifest values that correspond to ethical virtues and that are socially accepted, appreciated, and sought-after like health, sustainability, safety, etc.

Akin to or overlapping with machine biases in data sets are biases that stem from human-computer interactions. They arise from tendencies in human behavior when individuals interact with digital devices or platforms. These biases can emerge when individuals interact with each other while the interaction is mediated by the platforms themselves, or when they produce, evaluate, or seek particular content. Speaking from an ethical perspective, promoting human-computer interaction biases means putting individuals at a "disadvantage" who show a kind of human-computer interaction that originates in behavioral patterns that are socially less esteemed like detrimental norm violations, bad language, risky behavior, impulsiveness, incorrect beliefs, etc. On the other hand, one wants to promote friendliness, literacy, truthfulness, positive emotionality, prosocial orientation, etc.

### 4.2 Attraction poles in human and machine behavior

Typically, behavioral data are the result of tracking online activities of all kinds, meaning user-generated content, expressed or implicit relations between people, or behavioral traces (Olteanu et al. 2019). Different modes of behavior eventuate in different data contexts. Individuals leave different data traces behind depending on their emotional state, educational background, intelligence, wealth, age, moral maturity, and the like. In order to sort those traits and to classify human behavior and stages of development, one can draw on well-established theories in psychology and sociology. Within the framework of these theories, the aim is to distinguish different modes of behavior or stages of development according to empirical findings. As a general rule, behavior or personality development is understood to be largely a product of one's social environments. Those environments are classified, for instance, with the help of theories of social stratification (Grusky 2019; Vester 2001; Schulze 1996; Erikson et al. 1979; Bourdieu 1984). A person's milieu, meaning, simplistically speaking, upper, middle, or lower classes, determines their habitus, which in turn determines parts of their behavioral routines and vice versa. Individuals occupy a certain position in "social space" which is the result of a contested distribution of resources, meaning economic, cultural, social, or symbolic capital (Bourdieu 1989). The position an individual occupies in social space is in large parts "hereditary" and can be affected by social injustices. Nevertheless, the amount of capital a person can concentrate on her- or himself has a structuring power on many areas of life, meaning that it organizes a person's taste, language, estate, political orientation, or, to say it more generally, his or her dispositions.

Further, these dispositions also structure and have an impact on the way a person uses digital technologies, and influence what kind of data are tracked by these technologies. By using terms like "media-based inequalities", "digital divide" or "digital inequality", several studies show the strong influence a user's socioeconomic status has on media or Internet usage patterns (McCloud et al. 2016; Zillien and Hargittai 2009; boyd 2012; Hargittai 2008; Mossberger et al. 2003). Individuals with a higher socioeconomic status are more likely to engage in online activities that enhance their social position, have status-specific interests, interact more frequently with e.g. political or economic news or health information, have higher levels of computer literacy, use less often chat platforms or social networking sites, and so forth. All in all, the position of an individual in social space heavily influences his or her ways of using digital technologies and hence the kind of behavioral data that are digitally recorded – with the respective consequences for ethical quality dimensions of data sets.

While behavior is in many respects an outcome of the respective social environment, class, milieu, or social position, the same holds true for personality development, which is widely dependent on the circumstances of socialization. According to theories from developmental psychology, "higher" forms of personality development lead to other behavior patterns than "lower" ones (Baltes et al. 1978; Hart et al. 1997; Kohlberg et al. 1983). Normally, more cognitive-moral growth leads to more socially desirable or acceptable behavior. Philosophical theories about ideal moral acting, ranging from Kant's categorical imperative (Kant 1977), Habermas' discursive will-formation (Habermas 1987), or Rawls' theory of social contract (Rawls 1999), imply that individuals possess fully developed cognitive capacities. In this context, one can assume that personality or character development may strive towards the target values and rationality standards of these models. In this context, we cannot go into further details on that. Eventually, character dispositions as well as problem-solving abilities, emotional intelligence, cognitive development, prosocial behavior, educational status, mental health, etc. can be measured or tracked via digital technologies (Kosinski et al. 2013; Kosinski et al. 2014). Many of these assessment dimensions have clear attraction poles that can be defined by ethical theories. Machine biases that result from human-computer interactions aim at exactly these attraction poles.

### 4.3 Case studies

In the following, four case studies shall demonstrate how machine biases can help to support ethical machine behavior. The first two case studies correspond to representational as well as population biases, which are among the most prevalent types of machine biases (Olteanu et al. 2019; Mehrabi et al. 2019). The latter two case studies fall more into the category of human-interaction machine biases, meaning that they originate not in the way data sets are sampled but in the way humans produce social data. However, both areas merge in the sense that training stimuli are intentionally reduced and filtered according to ethical criteria.

#### 4.3.1 Case study 4: Representational bias on social media platforms

Representational bias emerges from the way one samples a population. Representational biases are a subset of machine biases that cause problems when generalizing beyond the training domain. Representational bias arises when machine learning models are trained on a dataset that favors certain representations over others. It is commonly seen as a problem since it compromises the external validity of an analysis. However, representational bias can also be an advantage when selecting for certain types of personality traits or character types that are for instance proxies for or connected to gender, educational backgrounds, urban populations, etc. A domain where representational biases could be introduced is recommendation systems on social media platforms. Currently, the main goal of these systems is to increase user engagement in order to increase the likelihood of advertisement contact (Hagendorff 2019b). This mechanism, however, causes a lot of problems, from addiction to the dissemination of fake news and extremist content, filter bubbles, the promotion of hate speech, and the like (Elhai et al. 2017; Kuss and Griffiths 2017; Vosoughi et al. 2018; Hagendorff 2021b). However, when taking social responsibility seriously, platforms should rearrange their objectives towards values of a vital and fair public discourse, truth, and information quality. This means to change the methods for algorithmic measurement and determination of information relevance. In order to achieve this, platforms have to foster representational biases in training data sets – for instance by favoring representations of rational, effortful, reflective interactions over impulsive interactions. Social media platforms can measure in which "mode" individuals operate with user interfaces. Affective, system-1 user behavior could be tracked by things like reaction or comment speed, scrolling or reading behavior, or the susceptibility to nudging techniques. Training data for recommendation systems can thus be scraped from contexts where user-generated data do represent system-2-humancomputer-interactions to a certain amount. This way, recommendation systems can be optimized on behavioral data that represents fewer impulsive reactions. Thereby, instead of negatively affecting public discourse by helping the automatic spreading of content that is most suited to cause emotional arousal and impulsive reactions, social media platforms could algorithmically disseminate content that is less "toxic" for public discourse. But this can only be achieved when embracing machine biases for ethical reasons.

#### 4.3.2 Case study 5: Population bias in e-commerce platforms

Population bias is the result of a mismatch between the target population of a service or platform and the user characteristics of the population represented in the dataset (Mehrabi et al. 2019). Again, this is seen as a problem since it decreases the external validity of a measurement. But population biases can actually have significant advantages, making them an ethical requirement of digital platforms or services. Here, we want to underpin this claim with an example, namely machine learning applications in online shopping. E-commerce platforms use various methods to promote purchasing behavior, for instance, shopping search engines, product recommendation systems, product reviews, dynamic pricing, cross-selling, customer analytics tools, conversion rate optimization, conversion funnels, varying payment options, specific user interface designs, and the like. Commonly, the main machine behavior objective is to maximize revenues. But via using population biases, machine behavior objectives can be altered in favor of values of sustainability or public health, for instance. How does that work? Online retailers can track and analyze all kinds of customer and purchasing data, allowing them for customer segmentation. Typically, differentiations for types of customers are made purely from a sales perspective, distinguishing between loyal, impulsive, novice, etc. customers. Notwithstanding that, customers can be segmented along criteria like health- or eco-consciousness by analyzing their product views, shopping behavior, product reviews, search terms, personality, socio-demographic factors, and the like. This way, platforms could introduce a population bias, using only data from health- as well as eco-conscious customers to train models for product recommendations, dynamic pricing, or ranking algorithms for their search engines, to name just three major setting options. Such measures could significantly foster the extent to which e-commerce platforms promote more sustainable and healthier consumer behavior. But without intentionally utilizing population biases, this will very likely not be possible.

### 4.3.3 Case study 6: Behavioral bias in self-driving cars

Behavioral bias emerges from the way humans interact with each other, or how they interact with digital devices and machines (Olteanu et al. 2019). We argue that behavioral bias is not something that should be eradicated from data sets. Quite the contrary, it can be vital for trustworthy AI systems. One example of this is selfdriving cars. They are supposed to guarantee as much safety as possible (Koopman and Wagner 2017). Avoiding crashes is paramount to advance their deployment. In order to achieve this, autonomous vehicles need to show safe machine behavior, meaning to comply with safe overtaking maneuver rules, following behavior, emergency stops, safe cornering, or line choice rules. To meet this goal, autonomous cars must generalize from past traffic situations to new ones. Thus, training data in the form of video recordings and other sensor data, as well as annotations for these data, are decisive for the vehicles' autopilot models to train. Since it is rather expensive to acquire enough annotation data, in some autonomous vehicles, labels are collected via measuring behavioral cues from human drivers, e.g., acceleration, deceleration, steering, etc. in manual mode or during autopilot disagreement (Eady 2019). These labels are then linked to the footage of the vehicles' surroundings. Moreover, driving behavior data can be combined with customer data and further data sources.

Subsequently, traffic psychologists can help machine learning practitioners establish tools to classify data that represent decent, safe driving behavior. How does that work? Individual characteristics like gender, age, driving experience, distraction, attention, reaction time, visual function, sensation seeking, impulsivity, etc. predict risky driving behavior (Anstey et al. 2005; Fergusson et al. 2003; Wayne and Miller 2018). According to accident statistics and empirical investigations, individuals who cause fatal as well as non-fatal car crashes tend to be male, of young age, have high levels of aggressiveness, sensation seeking, and impulsivity, as well as some other traits like lower levels of income, poor mental health status, higher levels of neuroticism, possibly raised blood alcohol concentration, lower driving experience, and show various forms of antisocial behavior or higher levels of social deviance (Abdoli et al. 2015; Čubranić-Dobrodolac et al. 2017; Vaughn et al. 2011; Wang et al. 2019; West and Hall 1997). Many of these traits can be digitally detected with a certain degree of accuracy. They can furthermore be combined with additional cues like engine speed, pedal pressure, improper following behavior, speaker volume, driver body posture, gestures, head movement, and verbal outbursts to assess or predict a driver's safety level. By following this path, severe behavioral biases become part of machine learning models in autonomous vehicles. Finally, data that is related to unwanted driving behavior can be downgraded or excluded, thus training data that is fed into the models that determine machine behavior during autopilot is exempt from it. Eventually, behavioral bias can be a necessary requirement for safety in autopilot training for autonomous vehicles.

### 4.3.4 Case study 7: Content production bias in language generation

Content production bias arises from structural, lexical, semantic, and syntactic differences in the contents generated by users (Olteanu et al. 2019). This bias can, for instance, be of importance in language generation. Chatbots as well as speech assistants of all kinds are supposed to produce appropriate, sufficiently eloquent language that does not violate social norms, discriminate against certain groups of people, or perpetuate biases that are incorporated into training data (all of which is especially precarious in open domain conversations) (Bolukbasi et al. 2016; Danaher 2018; Silvervarg et al. 2012; Sheng et al. 2019; West et al. 2019). Data sources for natural language generation can be text corpora of all kinds. The generation is based on finding statistical patterns in these corpora, meaning books, forum posts, news articles, communication data, websites, etc., which then allow a machine learning model, among other things, to predict the next word in a sentence based on previous words. Biases are a "natural" part of large text corpora and hence large language models (Bender et al. 2021). But instead of just incorporating common biases, one can assess creators and publishers of texts and hence text qualities. States and traits that can be tracked in order to assess text data quality may range from an author's educational background or occupation, intelligence, the characteristics of his or her keyboard strokes or display touching behavior (backspacing, etc.), potentially the time between writing and posting, the used publication platform, filtering intermediates, and review processes, to the language skills themselves. Particularly text data that does not stem from professional writers like journalists, book authors, scientists, etc., but from lay people is expected to be of lower quality. Text data that is not editorially controlled and therefore did not undergo any kind of review or filtering intermediate may be interspersed with orthographic mistakes, poor syntax, smaller word pools, slang, invectives, strong biases, etc. Quality data contexts are to be assessed in dependence on the respective purpose of an application for natural language generation. Texts from the public domain may be suited to improve a chatbot's realism, hence its ability to produce convincing, authentic, and human-like everyday language. On the other hand, these texts can be infiltrated with aggressive, discriminatory, or offensive phrases. To avoid these and other pitfalls, the selection of text corpora that are used to train conversational robots should not follow the bigger-is-better-approach like many commercially developed chatbots do. Instead, the selection of corpora should be intentionally biased towards narrowing it down to digital writings that underwent a firm quality check through publishers, peer reviews, or media agencies, that are embedded in a sophisticated web of citations or links, or that stem from individuals with high levels of language skills. Moreover, language proficiency can be determined by assessing the structure, continuity, errors, vocabulary richness, length of sentence, or changes made to a text. By using these selection criteria for content production, biases are purposefully implemented in natural language models. Content production biases thus are improving the quality of natural language generation.

## 5 Conclusion

With this paper, we aim at a reassessment of the value of machine biases as well as at a new acknowledgement of the value of high-level cognitive inductive priors for machine learning algorithms. Machine biases can be appraised as ethical as well as technical advantage for trustworthy and efficient algorithms instead of being regarded as the foundation of algorithmic discrimination and constraints to accuracy of machine outputs. Current machine learning techniques often focus on large training data sets, accuracy towards mean values of whole populations, stable application scenarios that are devoid of significant novelties, etc. Implementing ethical and cognitive machine biases, however, can help to effectively navigate uncertain, complex, and rapidly changing real-world environments by ensuring constant ethical machine behavior and enhancing accuracy in algorithmic decision making, respectively.

Our quest for cognitive machine bias implementation is motivated from a technical point of view, depicting potential steps for advancements in learning algorithms' performance. To that end, we use research on the ecological rationality of human cognitive biases as an inspiration to make the case for a transition from simply increasing the amount of computation and training data to a qualitative amendment of machine learning architectures. The proposed kinds of cognitive machine biases may, similar to human biases, be interpreted as systematic misconceptions, insensitivities to probabilities, or even errors, but in order to effectively navigate and interact with complex environments and to make accurate decisions in uncertain situations, those can become an important cornerstone. To be more precise, they may help to mitigate bias-variance dilemmas, avoid proneness to overfitting, simulate human decision strategies in domains where this is of importance, make models more explainable, utilize shortcuts for effective learning, etc.

Furthermore, via ethical machine biases, a more selective way of processing data for training sets can be established. Specifically, we argue for the promotion of biases in data sets that lead to a preferability of features that are desirable from an ethical point of view. Since machine behavior is the result of exposure to particular training stimuli, it is important to systematically select especially these training stimuli that reflect social or behavioral traits that are ethically desirable. Hence, ethically justified biases in training stimuli should be utilized to predetermine and shape a machine's behavior. Different qualities of human participation in model training can be distinguished, and there should always be a data bias towards "good" and legitimate human influences on machines. In order to identify data sources that reflect ethically sound behavior, particular states and traits of individuals can be used as criteria to select subpopulations that are deemed to be the most competent or morally versed group for a specific task. In social media platforms, this can mean introducing representational bias in order to improve the methods for algorithmic measurement of information relevance and to disseminate less content that is "toxic" for public discourse. Further, in e-commerce platforms, implementing population bias can promote healthy, sustainable purchasing behavior. Or, in self-driving cars, behavioral bias can be used to foster safe autopilot behavior. And in natural language generation, content production bias can guide text corpus selection to improve the quality of natural language generation applications. All in all, instead of manipulating algorithms in order to suppress certain training stimuli, we argue for data set manipulation, so that the root cause of unethical machine behavior is addressed and not just the symptoms.

To sum up, we believe that the many arguments that can be made in favor of cognitive and ethical machine bias implementation add up to a strong trend calling for a re-evaluation and stronger acknowledgement of the significance of machine biases. This paper is a first and tentative step in this direction. We hope that in future research, cognitive scientists, machine learning researchers, as well as domain ethicists will work closely together to figure out in greater detail the potentials of a new, ethically motivated filter regime for the selection of biased training data as well as the potentials of the structure-wise implementation of human cognitive biases into learning algorithms.

## Acknowledgements
This research was supported by the Cluster of Excellence "Machine Learning – New Perspectives for Science" funded by the Deutsche Forschungsgemeinschaft (DFG, German Research Foundation) under Germany's Excellence Strategy – Reference Number EXC 2064/1 – Project ID 390727645. We would like to thank Peter Dayan and Felix Wichmann for helpful comments on the manuscript.

## Author contribution
Both Sarah Fabi and Thilo Hagendorff equally developed and discussed the ideas for the paper and contributed to the manuscript.